\documentclass[11pt]{article}
\usepackage{coling2018}
\usepackage{times}
\usepackage{url}
\usepackage{latexsym}
\usepackage{graphicx}

\usepackage{amsmath}
\usepackage{amssymb}

\title{UnibucKernel Reloaded: First Place in Arabic Dialect Identification for the Second Year in a Row}

\author{Andrei M. Butnaru \and Radu Tudor Ionescu\\
  Department of Computer Science, University of Bucharest\\
  14 Academiei, Bucharest, Romania\\
  {\tt butnaruandreimadalin@gmail.com, raducu.ionescu@gmail.com}\\
}

\date{}

\begin{document}
\maketitle
\vspace*{-0.3cm}
\begin{abstract}
\vspace*{-0.3cm}
We present a machine learning approach that ranked on the first place in the Arabic Dialect Identification (ADI) Closed Shared Tasks of the 2018 VarDial Evaluation Campaign. The proposed approach combines several kernels using multiple kernel learning. While most of our kernels are based on character $p$-grams (also known as $n$-grams) extracted from speech or phonetic transcripts, we also use a kernel based on dialectal embeddings generated from audio recordings by the organizers. In the learning stage, we independently employ Kernel Discriminant Analysis (KDA) and Kernel Ridge Regression (KRR). Preliminary experiments indicate that KRR provides better classification results. Our approach is shallow and simple, but the empirical results obtained in the 2018 ADI Closed Shared Task prove that it achieves the best performance. Furthermore, our top macro-$F_1$ score ($58.92\%$) is significantly better than the second best score ($57.59\%$) in the 2018 ADI Shared Task, according to the statistical significance test performed by the organizers. Nevertheless, we obtain even better post-competition results (a macro-$F_1$ score of $62.28\%$) using the audio embeddings released by the organizers after the competition. With a very similar approach (that did not include phonetic features), we also ranked first in the ADI Closed Shared Tasks of the 2017 VarDial Evaluation Campaign, surpassing the second best method by $4.62\%$. We therefore conclude that our multiple kernel learning method is the best approach to date for Arabic dialect identification.
\end{abstract}

\setlength{\abovedisplayskip}{3pt}
\setlength{\belowdisplayskip}{3pt}

\blfootnote{This work is licensed under a Creative Commons Attribution 4.0 International License. License details: \url{http://creativecommons.org/licenses/by/4.0/}.}

\section{Introduction}
\vspace*{-0.2em}


The 2016 and 2017 VarDial Evaluation Campaigns~\cite{dsl2016,dsl2017} indicate that dialect identification is a challenging NLP task, actively studied by researchers in nowadays. Based solely on speech transcripts, the top two Arabic dialect identification (ADI) systems~\cite{Ionescu-VarDial-2016,Malmasi-ADI-2016} that participated in the 2016 ADI Shared Task~\cite{dsl2016} attained weighted $F_1$ scores just over $50\%$, in a 5-way classification setting. For the 2017 ADI Shared Task~\cite{dsl2017}, the organizers provided audio features along with speech transcripts. The top two systems~\cite{Ionescu-VarDial-2017,malmasi-zampieri:2017:VarDial2} were able to reach weighted $F_1$ scores above $70\%$ by using audio features and by including the samples from the development set into the training set. For the 2018 ADI Shared Task~\cite{vardial2018report}, the organizers have added phonetic features along with audio features and speech transcripts. To this end, we present our approach to the 2018 ADI Shared Task, which is based on adding string kernels computed on phonetic transcripts to the multiple kernel learning model~\cite{Ionescu-VarDial-2017} that we previously designed for the 2017 ADI Shared Task. In the 2018 ADI Shared Task, the participants had to discriminate between Modern Standard Arabic (MSA) and four Arabic dialects, in a 5-way classification setting. A number of $6$ teams have submitted their results on the test set, and our team (UnibucKernel) ranked on the first place with an accuracy of $58.65\%$ and a macro-$F_1$ score of $58.92\%$. 

Our best scoring system in the ADI Shared Task combines several kernels using multiple kernel learning. The first kernel that we considered is the $p$-grams presence bits kernel\footnote{We computed the $p$-grams presence bits kernel using the code available at http://string-kernels.herokuapp.com.}, which takes into account only the presence of $p$-grams instead of their frequency. The second kernel is the (histogram) intersection string kernel\footnote{We computed the intersection string kernel using the code available at http://string-kernels.herokuapp.com.}, which was first used in a text mining task by \newcite{Ionescu-EMNLP-2014}. The third kernel is derived from Local Rank Distance (LRD)\footnote{We computed the Local Rank Distance using the code available at http://lrd.herokuapp.com.}, a distance measure that was first introduced in computational biology~\cite{Ionescu-SYNASC-2013,Ionescu-PONE-2014}, but it has also shown its application in NLP~\cite{Popescu-BEA8-2013,Ionescu-ICONIP-2015b}. These string kernels have been previously used for Arabic dialect identification from speech transcripts by~\newcite{Ionescu-VarDial-2016}, and they obtained very good results, taking the second place in the 2016 ADI Shared Task~\cite{dsl2016}. In this paper, we apply string kernels on speech transcripts as well as phonetic transcripts, obtaining better results. While most of our kernels are based on character $p$-grams from speech or phonetic transcripts, we also use an RBF kernel~\cite{taylor-Cristianini-cup-2004} based on dialectal embeddings automatically generated from audio recordings using the approach described in~\cite{Shon-Odyssey-2018}. In our previous work~\cite{Ionescu-VarDial-2017}, we have successfully combined string kernels computed on speech transcripts with an RBF kernel computed on audio features and we ranked on the first place in the 2017 ADI Shared Task~\cite{dsl2017}. For the 2018 ADI Shared Task, our multiple kernel learning method includes string kernels computed on phonetic transcripts along with the other kernels.

We considered two kernel classifiers~\cite{taylor-Cristianini-cup-2004} for the learning task, namely Kernel Ridge Regression (KRR) and Kernel Discriminant Analysis (KDA). 
In a set of preliminary experiments performed on the ADI development set, we found that KRR gives slightly better results than KDA. In the end, we decided to submit results using only KRR.
Before submitting our results, we have also tuned our string kernels for the task. First of all, we tried out $p$-grams of various lengths, including blended variants of string kernels as well. Second of all, we evaluated the individual kernels and various kernel combinations. The empirical results indicate that string kernels computed on speech transcripts attain significantly better performance than string kernels computed on phonetic transcripts. When we combined the string kernels computed on speech transcripts with the kernel base on audio embeddings, we found that the performance improves by nearly $5\%$. We obtained another improvement of almost $1\%$ when we included the string kernels computed on phonetic transcripts. 
All these choices played an important role in obtaining the first place in the final ranking of the 2018 ADI Shared Task.

The paper is organized as follows. Work related to Arabic dialect identification and to methods based on string kernels is presented in Section~\ref{sec_Related_Work}. Section~\ref{sec_Similarity_Measures} presents the kernels that we used in our approach. The learning methods employed in the experiments are described in Section~\ref{sec_Learning_Methods}. Details about the Arabic dialect identification experiments are provided in Sections~\ref{sec_ADI_Experiments}. Finally, we draw our conclusion in Section~\ref{sec_Conclusion}.

\vspace*{-0.2em}
\section{Related Work}
\label{sec_Related_Work}
\vspace*{-0.2em}
\subsection{Arabic Dialect Identification}

Arabic dialect identification is a relatively new NLP task with only a handful of works~\cite{Biadsy-2009,Zaidan-2011,Elfardy-ACL-2013,Darwish-EMNLP-2014,Zaidan-COLI-2014,malmasi-et-al:2015:adi} to address it before the 2016 VarDial Evaluation Campaign~\cite{dsl2016}. Although it did not received too much attention before 2016, the task is very important for Arabic NLP tools, as most of these tools have only been design for Modern Standard Arabic. \newcite{Biadsy-2009} describe a phonotactic approach that automatically identifies the Arabic dialect of a speaker given a sample of speech. While \newcite{Biadsy-2009} focus on spoken Arabic dialect identification, others have tried to identify the Arabic dialect of given texts~\cite{Zaidan-2011,Elfardy-ACL-2013,Darwish-EMNLP-2014,malmasi-et-al:2015:adi}. \newcite{Zaidan-2011} introduce the Arabic Online Commentary (AOC) data set of 108K labeled sentences, $41\%$ of them having dialectal content. They employ a language model for automatic dialect identification on their collected data. A supervised approach for sentence-level dialect identification between Egyptian and MSA is proposed by \newcite{Elfardy-ACL-2013}. Their system outperforms the approach presented by \newcite{Zaidan-2011} on the same data set. \newcite{Zaidan-COLI-2014} extend their previous work~\cite{Zaidan-2011} and conduct several ADI experiments using word and character $p$-grams. Different from most of the previous work, \newcite{Darwish-EMNLP-2014} have found that word unigram models do not generalize well to unseen topics. They suggest that lexical, morphological and phonological features can capture more relevant information for discriminating dialects. As the AOC corpus is not controlled for topical bias, \newcite{malmasi-et-al:2015:adi} also state that the models trained on this corpus may not generalize to other data as they implicitly capture topical cues. They perform ADI experiments on the Multidialectal Parallel Corpus of Arabic (MPCA)~\cite{Bouamor-LREC-2014} using various word and character $p$-grams models in order to assess the influence of topical bias. Interestingly, \newcite{malmasi-et-al:2015:adi} find that character $p$-grams are ``in most scenarios the best single feature for this task'', even in a cross-corpus setting. Their findings are consistent with the results of~\newcite{Ionescu-VarDial-2016} in the 2016 ADI Shared Task~\cite{dsl2016}, as~\newcite{Ionescu-VarDial-2016} ranked on the second place using solely character $p$-grams from Automatic Speech Recognition (ASR) transcripts. The 2017 ADI Shared Task data set~\cite{Ali-2016} contains the original audio files and some low-level audio features, called i-vectors, along with the ASR transcripts of Arabic speech collected from the Broadcast News domain. The results of the top systems~\cite{Ionescu-VarDial-2017,malmasi-zampieri:2017:VarDial2} in the 2017 ADI Shared Task~\cite{dsl2017} indicate that the audio features produce a much better performance, probably because there are many ASR errors (perhaps more in the dialectal speech segments) that make Arabic dialect identification from ASR transcripts much more difficult. In 2017, ADI has attracted a higher attention in the scientific community, since researchers have organized three independent shared tasks~\cite{ali2017speech,rangel2017overview,dsl2017}, with $32$ participants in total. For a comprehensive and complete survey on ADI, we refer the reader to the work of \cite{jauhiainen2018automatic}.

\vspace*{-0.2em}
\subsection{String Kernels}

From the beginning of the 21st century to the present days, methods of handling text at the character level have demonstrated impressive performance levels in various text analysis tasks~\cite{LodhiSSCW02,Sanderson:06,Kate-ACL-2006,Escalante-ACL-2011,PopescuG12,Ionescu-EMNLP-2014,Ionescu-COLI-2016,franco-EACL-2017,Popescu-KES-2017,Cozma-ACL-2018}. String kernels are a common form of using information at the character level. They are a particular case of the more general convolution kernels~\cite{haussler-1999}. \newcite{LodhiSSCW02} used string kernels for document categorization with very good results. String kernels were also successfully used in authorship identification~\cite{Sanderson:06,PopescuG12}. For example, the system described by~\newcite{PopescuG12} ranked first in most problems and overall in the PAN 2012 Traditional Authorship Attribution tasks. More recently, various blended string kernels reached state-of-the-art accuracy rates for native language identification~\cite{Ionescu-COLI-2016,Ionescu-BEA-2017}, Arabic dialect identification~\cite{Ionescu-VarDial-2016,Ionescu-VarDial-2017}, polarity classification~\cite{franco-EACL-2017,Popescu-KES-2017} and automatic essay scoring~\cite{Cozma-ACL-2018}. 

\vspace*{-0.2em}
\section{Kernels for Arabic Dialect Identification}
\label{sec_Similarity_Measures}
\vspace*{-0.2em}

\subsection{String Kernels}
\label{sec_String_Kernels}

Kernel functions \cite{taylor-Cristianini-cup-2004} capture the intuitive notion of similarity between objects in a specific domain. For strings, many such kernel functions exist with various applications in computational biology and computational linguistics. String kernels embed the texts in a very large feature space, given by all the substrings of length $p$, and leave the job of selecting important (discriminative) features for the specific classification task to the learning algorithm, which assigns higher weights to the important features (character $p$-grams). Perhaps one of the most natural ways to measure the similarity of two strings is to count how many substrings of length $p$ the two strings have in common. This gives rise to the $p$-spectrum kernel. Formally, for two strings over an alphabet $\Sigma$, $s,t \in \Sigma^*$, the $p$-spectrum kernel is defined as:
\begin{equation*}
\begin{split}
k_p(s,t)=\sum\limits_{v \in \Sigma^p}\mbox{num}_v(s) \cdot \mbox{num}_v(t),
\end{split}
\end{equation*}
where $\mbox{num}_v(s)$ is the number of occurrences of string $v$ as a substring in $s$.\footnote{The notion of substring requires contiguity. \newcite{taylor-Cristianini-cup-2004} discuss the ambiguity between the terms \emph{substring} and \emph{subsequence} across different domains: biology, computer science.}
The feature map defined by this kernel associates to each string a vector of dimension $|\Sigma|^p$ containing the histogram of frequencies of all its substrings of length $p$ ($p$-grams).
A variant of this kernel can be obtained if the embedding feature map is modified to associate to each string a vector of dimension $|\Sigma|^p$ containing the presence bits (instead of frequencies) of all its substrings of length $p$. Thus, the character $p$-grams presence bits kernel is obtained:
\begin{equation*}
\begin{split}
k^{0/1}_p(s,t)=\sum\limits_{v \in \Sigma^p}\mbox{in}_v(s) \cdot \mbox{in}_v(t),
\end{split}
\end{equation*}
where $\mbox{in}_v(s)$ is $1$ if string $v$ occurs as a substring in $s$, and $0$ otherwise.

In computer vision, the (histogram) intersection kernel has successfully been used for object class recognition from images~\cite{DBLP:conf/cvpr/MajiBM08,Vedaldi-add-ker-2010}. \newcite{Ionescu-EMNLP-2014} have used the intersection kernel as a kernel for strings, in the context of native language identification. The intersection string kernel is defined as follows:
\begin{equation*}
\begin{split}
k^{\cap}_p(s,t)=\sum\limits_{v \in \Sigma^p} \min \lbrace \mbox{num}_v(s), \mbox{num}_v(t) \rbrace .
\end{split}
\end{equation*}

For the $p$-spectrum kernel, the frequency of a $p$-gram has a very significant contribution to the kernel, since it considers the product of such frequencies. On the other hand, the frequency of a $p$-gram is completely disregarded in the $p$-grams presence bits kernel. The intersection kernel lies somewhere in the middle between the $p$-grams presence bits kernel and the $p$-spectrum kernel, in the sense that the frequency of a $p$-gram has a moderate contribution to the intersection kernel. In other words, the intersection kernel assigns a high score to a $p$-gram only if it has a high frequency in both strings, since it considers the minimum of the two frequencies. The $p$-spectrum kernel assigns a high score even when the $p$-gram has a high frequency in only one of the two strings. Thus, the intersection kernel captures something more about the correlation between the $p$-gram frequencies in the two strings. Based on these comments, we decided to use only the $p$-grams presence bits kernel and the intersection string kernel in the ADI experiments.

Data normalization helps to improve machine learning performance for various applications. Since the value range of raw data can have large variations, classifier objective functions will not work properly without normalization. 
After normalization, each feature has an approximately equal contribution to the similarity between two samples.
%
To obtain a normalized kernel matrix of pairwise similarities between samples, each component is divided by the square root of the product of the two corresponding diagonal components:
\begin{equation*}\label{eq_Kernel_Matrix_Normalization}
\begin{split}
\hat{K}_{ij} = \frac{K_{ij}}{\sqrt{K_{ii} \cdot K_{jj}}}.
\end{split}
\end{equation*}
To ensure a fair comparison among strings of different lengths, we use normalized versions of the $p$-grams presence bits kernel and the intersection kernel in our experiments.
Taking into account $p$-grams of different lengths and summing up the corresponding kernels,
new kernels, termed \emph{blended spectrum kernels}, can be obtained. We have used various blended spectrum kernels in the ADI experiments in order to find the best combination. It is important to mention that we applied the blended $p$-grams presence bits kernel and the blended intersection kernel on both speech transcripts and phonetic transcripts.


\vspace*{-0.2em}
\subsection{Kernel based on Local Rank Distance}
\label{sec_LRD}

Local Rank Distance~\cite{Ionescu-SYNASC-2013} is a recently introduced distance that measures the non-alignment score between two strings. It has already shown promising results in computational biology~\cite{Ionescu-SYNASC-2013,Ionescu-PONE-2014} and native language identification~\cite{Popescu-BEA8-2013,Ionescu-ICONIP-2015b}.

In order to describe LRD, we use the following notations. Given a string $x$ over an alphabet $\Sigma$, 
the length of $x$ is denoted by $|x|$. 
Strings are considered to be indexed starting from position $1$, that is $x = x[1]x[2]\cdots x[|x|]$. Moreover, $x[i:j]$ denotes its substring $x[i]x[i+1]\cdots x[j-1]$. Given a fixed integer $p \geq 1$, a threshold $m \geq 1$, and two strings $x$ and $y$ over $\Sigma$, the \emph{Local Rank Distance} between $x$ and $y$, denoted by $\Delta_{LRD}(x,y)$, is defined through the following algorithmic process. For each position $i$ in $x$ ($1 \leq i \leq |x|- p + 1$), the algorithm searches for that position $j$ in $y$ ($1 \leq j \leq |y| - p + 1$) such that $x[i:i+p] = y[j:j+p]$ and $|i - j|$ is minimized. If $j$ exists and $|i - j| < m$, then the offset $|i - j|$ is added to the Local Rank Distance. Otherwise, the maximal offset $m$ is added to the Local Rank Distance. 
LRD is focused on the local phenomenon, and tries to pair identical $p$-grams at a minimum offset. To ensure that LRD is a (symmetric) distance function, the algorithm also has to sum up the offsets obtained from the above process by exchanging $x$ and $y$. LRD is formally defined in~\cite{Ionescu-SYNASC-2013,Ionescu-PONE-2014,Ionescu-ACVPR-2016}.


The search for matching $p$-grams is limited within a window of fixed size. The size of this window is determined by the maximum offset parameter $m$. We set $m = 300$ in our experiments, which is larger than the maximum length of the ASR transcripts provided in the training set. 
In the experiments, the efficient algorithm of \newcite{Ionescu-ICONIP-2015b} is used to compute LRD.
However, LRD needs to be used as a kernel function. We use the RBF kernel~\cite{taylor-Cristianini-cup-2004} to transform LRD into a similarity measure:
\begingroup
\large
\begin{equation*}
\hat{k}^{LRD}_p(s, t) = exp \left(- \frac{\displaystyle \Delta_{LRD}(s, t)} {\displaystyle 2 \sigma^2} \right),
\end{equation*}
\endgroup
where $s$ and $t$ are two strings and $p$ is the $p$-grams length. The parameter $\sigma$ is usually chosen
so that values of $\hat{k}(s, t)$ are well scaled. We have tuned $\sigma$ in a set of preliminary experiments. In the above equation, $\Delta_{LRD}$ is already normalized to a value in the $[0,1]$ interval to ensure a fair comparison of strings of different length. The resulted similarity matrix is then squared, i.e. $K = K \cdot K'$, to ensure that it becomes a symmetric and positive definite kernel matrix. Due to time constraints, we applied the kernel based on Local Rank Distance only on speech transcripts.

\vspace*{-0.2em}
\subsection{Kernel based on Audio Features}

Along with the string kernels, we also build a kernel from the audio embeddings provided with the data set~\cite{Shon-Odyssey-2018}. The dialectal embeddings are generated by training convolutional neural networks on audio recordings, as described in~\cite{Shon-Odyssey-2018,Najafian-ICASSP-2018}. The provided embeddings have $600$ dimensions. In order to build a kernel from the audio embeddings, we first compute the Euclidean distance between each pair of embedding vectors. We then employ the RBF kernel to transform the distance into a similarity measure:
\begingroup
\large
\begin{equation*}
\hat{k}_{\textit{audio}}(x, y) = exp \left(- \frac{ \sqrt{\sum^{m}_{j=1}(x_j - y_j)^2}} { 2 \sigma^2} \right),
\end{equation*}
\endgroup
where $x$ and $y$ are two audio embedding vectors and $m$ represents the size of the two embedding vectors, $600$ in our case. For optimal results, we have tuned the parameter $\sigma$ in a set of preliminary experiments. As for the LRD kernel, we square the kernel matrix of the kernel based on audio embeddings.

\section{Learning Methods}
\label{sec_Learning_Methods}

Kernel-based learning algorithms work by embedding the data into a Hilbert feature space and by searching for linear relations in that space. The embedding is performed implicitly, by specifying the inner product between each pair of points rather than by giving their coordinates explicitly. More precisely, a kernel matrix that contains the pairwise similarities between every pair of training samples is used in the learning stage to assign a vector of weights to the training samples. Let $\alpha$ denote this weight vector. In the test stage, the pairwise similarities between a test sample $x$ and all the training samples are computed. Then, the following binary classification function assigns a positive or a negative label to the test sample:
\begin{equation*}\label{eq_dual_decision}
\begin{split}
g(x) = \sum_{i=1}^n \alpha_i \cdot k(x, x_i),
\end{split}
\end{equation*}
where $x$ is the test sample, $n$ is the number of training samples, $X = \lbrace x_1, x_2, ..., x_n \rbrace$ is the set of training samples, $k$ is a kernel function, and $\alpha_i$ is the weight assigned to the training sample $x_i$. 

The advantage of using the dual representation induced by the kernel function becomes clear if the dimension of the feature space $m$ is taken into consideration. Since string kernels are based on character $p$-grams, the feature space is indeed very high. For instance, using $5$-grams based only on the $28$ letters of the basic Arabic alphabet will result in a feature space of $28^5 = 17,210,368$ features. However, our best kernels are based on a feature space that includes $3$-grams, $4$-grams, $5$-grams and $6$-grams. As long as the number of samples $n$ is much lower than the number of features $m$, it can be more efficient to use the dual representation given by the kernel matrix. This fact is also known as the \emph{kernel trick}~\cite{taylor-Cristianini-cup-2004}.

Various kernel methods differ in the way they learn to separate the samples. In the case of binary classification problems, kernel-based learning algorithms look for a discriminant function, a function that assigns $+1$ to examples belonging to one class and $-1$ to examples belonging to the other class. In the ADI experiments, we used the Kernel Ridge Regression (KRR) binary classifier. Kernel Ridge Regression selects the vector of weights that simultaneously has small empirical error and small norm in the Reproducing Kernel Hilbert Space generated by the kernel function. 
KRR is a binary classifier, but Arabic dialect identification is a multi-class classification problem. There are many approaches for combining binary classifiers to solve multi-class problems. Typically, the multi-class problem is broken down into multiple binary classification problems using common decomposition schemes such as: one-versus-all and one-versus-one. We considered the one-versus-all scheme for our Arabic dialect classification task. There are also kernel methods that take the multi-class nature of the problem directly into account, for instance Kernel Discriminant Analysis. The KDA classifier is sometimes able to improve accuracy by avoiding the masking problem~\cite{ESL-hastie-tibshirani-2003}. 
More details about KRR and KDA are given in~\cite{taylor-Cristianini-cup-2004}.

\section{Experiments on Arabic Dialects}
\label{sec_ADI_Experiments}

\subsection{Data Set}

The 2018 ADI Shared Task data set~\cite{Ali-2016} contains audio recordings, ASR transcripts and phonetic transcripts of Arabic speech collected from the Broadcast News domain. The task is to discriminate between Modern Standard Arabic (MSA) and four Arabic dialects, namely Egyptian (EGY), Gulf (GLF), Levantine (LAV), and North-African or Maghrebi (NOR). Although the data set is similar to those used in the 2016 and the 2017 ADI Shared Tasks~\cite{dsl2016,dsl2017}, this year the organizers provided phonetic transcripts produced by four non-Arabic automatic phoneme recognizers (Czech, English, Hungarian and Russian), which perform long temporal phoneme recognition and have been previously shown to be useful for discriminating between the dialects of various languages.

\subsection{Parameter and System Choices}

In our approach, we treat both ASR transcripts and phonetic transcripts as strings. Because the approach works at the character level, there is no need to split the texts into tokens, or to do any NLP-specific processing before computing the string kernels. The only editing done to the transcripts was the replacing of sequences of consecutive space characters (space, tab, and so on) with a single space character. This normalization was needed in order to prevent the artificial increase or decrease of the similarity between texts, as a result of different spacing.

\begin{table}
\center
\begin{tabular}{|l|l|c|c|c|}
\hline
\bf Kernel 			& \bf Input 	                    & \bf Range of $\mathbf{p}$-grams& \bf Accuracy  & \bf Macro-$\mathbf{F_1}$\\
\hline
\hline
$\hat{k}^{0/1}$ 	& Speech transcripts                & $3$-$6$                   & $54.85\%$ & $53.94\%$ \\
$\hat{k}^{\cap}$ 	& Speech transcripts                & $3$-$6$                   & $54.53\%$ & $53.77\%$ \\
$\hat{k}^{LRD}$     & Speech transcripts                & $3$-$6$                   & $51.09\%$ & $50.15\%$ \\
\hline
$\hat{k}^{0/1}$ 	& Czech phonetic transcripts        & $3$-$6$                   & $39.14\%$ & $39.43\%$ \\
$\hat{k}^{\cap}$ 	& Czech phonetic transcripts        & $3$-$4$                   & $39.08\%$ & $38.52\%$ \\
$\hat{k}^{0/1}$ 	& English phonetic transcripts      & $9$-$10$                  & $31.42\%$ & $30.49\%$ \\
$\hat{k}^{\cap}$ 	& English phonetic transcripts      & $9$-$11$                  & $31.36\%$ & $30.47\%$ \\
$\hat{k}^{0/1}$ 	& Hungarian phonetic transcripts    & $3$-$5$                   & $39.78\%$ & $39.40\%$ \\
$\hat{k}^{\cap}$ 	& Hungarian phonetic transcripts    & $3$-$5$                   & $40.36\%$ & $40.39\%$ \\
$\hat{k}^{0/1}$ 	& Russian phonetic transcripts      & $3$-$5$                   & $36.08\%$ & $36.28\%$ \\
$\hat{k}^{\cap}$ 	& Russian phonetic transcripts      & $3$-$4$                   & $38.19\%$ & $38.04\%$ \\
\hline
\end{tabular}
\caption{Optimal ranges of $p$-grams and corresponding results for each string kernel and each input type on the development set of the ADI Shared Task. All results are obtained using KRR as classifier.}
\label{tab_pgrams}
\end{table}

In order to tune the parameters and find the best system choices, we used the development set. We first carried out a set of preliminary experiments to determine the optimal range of $p$-grams for each string kernel and each type of input, i.e. speech transcripts or phonetic transcripts in four languages. We fixed the learning method to KRR and we evaluated all the $p$-grams in the range $2$-$12$. The optimal range of $p$-grams and the associated accuracy and macro-$F_1$ score for each kernel and each input is presented in Table~\ref{tab_pgrams}. In most cases, it seems that the optimal range of $p$-grams is $3$-$5$ or $3$-$6$. An exception is the range of $p$-grams that provides the best results on English phonetic transcripts. Regarding the input type, it is clear that string kernels provide better results when they are applied on speech transcripts. With an accuracy of $54.85\%$ and a macro-$F_1$ score of $53.94\%$, the best individual kernel on the development set is the blended presence bits kernel ($\hat{k}^{0/1}$) computed on speech transcripts. On the other hand, the best kernel computed on (Hungarian) phonetic transcripts is the blended intersection kernel ($\hat{k}^{\cap}$), which obtains an accuracy of $40.36\%$ and a macro-$F_1$ score of $40.39\%$.

\begin{table}
\center
\begin{tabular}{|l|l|c|c|}
\hline
\bf Kernels 			                                            & \bf Input & \bf Accuracy  & \bf Macro-$\mathbf{F_1}$\\
\hline
\hline
$\hat{k}_{speech} = \hat{k}^{0/1}_{3-6} + \hat{k}^{\cap}_{3-6} + \hat{k}^{LRD}_{3-6}$	& Speech transcripts    & $55.17\%$ & $54.72\%$ \\
\hline
$\hat{k}_{CZ} = \hat{k}^{0/1}_{3-6} + \hat{k}^{\cap}_{3-4}$         & Czech phonetic transcripts                & $39.72\%$ & $39.71\%$ \\
$\hat{k}_{EN} = \hat{k}^{0/1}_{9-10} + \hat{k}^{\cap}_{9-11}$ 	    & English phonetic transcripts              & $31.42\%$ & $30.54\%$ \\
$\hat{k}_{HU} = \hat{k}^{0/1}_{3-5} + \hat{k}^{\cap}_{3-5}$ 	    & Hungarian phonetic transcripts            & $41.06\%$ & $40.85\%$ \\
$\hat{k}_{RU} = \hat{k}^{0/1}_{3-5} + \hat{k}^{\cap}_{3-4}$ 	    & Russian phonetic transcripts              & $38.38\%$ & $37.75\%$ \\
$\hat{k}_{phonetic} = \hat{k}_{CZ} + \hat{k}_{EN} + \hat{k}_{HU} + \hat{k}_{RU}$    & All phonetic transcripts  & $45.02\%$ & $45.06\%$ \\
\hline
$\hat{k}_{\textit{audio}}$                                          & Audio embeddings                & $52.87\%$ & $52.72\%$ \\
\hline
\end{tabular}
\caption{Results of best kernel combinations for each input type on the development set of the ADI Shared Task. All results are obtained using KRR as classifier.}
\label{tab_comb_per_input}
\vspace*{-0.5em}
\end{table}

After determining the optimal range of $p$-grams for each kernel and input pair, we conducted further experiments by combining the kernels for each type of input. When multiple kernels are combined, the features are actually embedded in a higher-dimensional space. As a consequence, the search space of linear patterns grows, which helps the classifier to select a better discriminant function. We adopt the most natural way of combining two or more kernels, namely we simply sum up the corresponding kernels. The process of summing up kernels or kernel matrices is equivalent to feature vector concatenation. The results of kernel combinations for each input type are presented in Table~\ref{tab_comb_per_input}. For the phonetic transcripts, we tried out kernel combinations for each non-Arabic phoneme recognizer, as well as a kernel combination on all phonetic transcripts. Compared to the individual components, we observed the highest improvement when we combined all string kernels based on phonetic transcripts. Indeed, the best combination (presented in Table~\ref{tab_comb_per_input}) using only phonetic transcripts as input reaches an accuracy of $45.02\%$ and a macro-$F_1$ score of $45.06\%$, while the best individual kernel (presented in Table~\ref{tab_pgrams}) applied over phonetic transcripts reaches an accuracy of $40.36\%$ and a macro-$F_1$ score of $40.39\%$. For the speech transcripts, the kernel combination is better than each individual component, but the performance gain is not as high as in the case of phonetic transcripts. For the audio recordings, we used only a single kernel based on audio embeddings ($\hat{k}_{\textit{audio}}$). We tuned the parameter $\sigma$ of the RBF kernel based on audio embeddings, and the best option seems to be $\sigma = 1$, which produces an accuracy of $52.87\%$ and a macro-$F_1$ score of $52.72\%$.

\begin{table}
\center
\begin{tabular}{|l|l|c|c|c|c|}
\hline
\bf Kernels 			                                            & \multicolumn{2}{|c|}{\bf KRR}              & \multicolumn{2}{|c|}{\bf KDA}\\
\cline{2-5}
 			                                            & \bf Accuracy  & \bf Macro-$\mathbf{F_1}$   & \bf Accuracy  & \bf Macro-$\mathbf{F_1}$\\
\hline
\hline
$\hat{k}_{speech} + \hat{k}_{\textit{audio}}$~\cite{Ionescu-VarDial-2017}                       & $59.52\%$         & $59.12\%$         & $58.05\%$         & $58.13\%$\\
$\hat{k}_{speech} + \hat{k}_{\textit{audio}} + \hat{k}_{phonetic}$  & $60.66\%$         & $60.46\%$         & $58.94\%$         & $59.20\%$\\
\hline
\end{tabular}
\caption{Accuracy rates and macro-$F_1$ scores of various kernels combined across different input types: audio recordings, speech transcripts and phonetic transcripts. Both KRR and KDA are alternatively employed for the learning task. The results are obtained on the ADI development set.}
\label{tab_dev_results}
\end{table}

Since we obtained different kernel representations from speech transcripts, phonetic transcripts and audio recordings, a good approach for improving the performance is to further combine the best kernel combinations in order to obtain a multiple kernel learning approach that benefits from all three types of input formats (audio embeddings, speech transcripts and phonetic transcripts). The results of our multiple kernel learning method on the development set are reported in Table~\ref{tab_dev_results}. It is important to note that, in Table~\ref{tab_dev_results}, we included the results of the multiple kernel learning approach~\cite{Ionescu-VarDial-2017} that ranked first the 2017 ADI Shared Task, as reference. We notice that the string kernels computed on phonetic features help to improve the performance over the last year's winning approach by almost $1\%$. We also note that the combination of string kernels computed on speech transcripts ($\hat{k}_{speech}$) and the kernel based on audio embeddings ($\hat{k}_{\textit{audio}}$) improves the accuracy of the individual components by more than $4\%$. Indeed, the kernel combination that includes $\hat{k}_{speech}$ and $\hat{k}_{\textit{audio}}$ attains an accuracy of $59.52\%$ and a macro-$F_1$ score of $59.12\%$ when KRR is employed for the learning task, while the better kernel component ($\hat{k}_{speech}$) reaches an accuracy of $55.17\%$ and a macro-$F_1$ score of $54.72\%$. Among the two kernel classifiers, it seems that KRR attains slightly better results than KDA. In fact, KRR is the only classifier that surpasses the $60\%$ performance threshold on the development set. This happens when the string kernels computed on phonetic features ($\hat{k}_{phonetic}$) are included in the multiple kernel learning framework. We note that, in all the experiments performed on the development set, KRR yields the best results when we use a regularization parameter of $10^{-3}$. In the end, we decided to submit three runs for final evaluation on the test set. All submissions are based on the KRR classifier. The first submission (run 1) is based on the sum of $\hat{k}_{speech}$, $\hat{k}_{\textit{audio}}$ and $\hat{k}_{phonetic}$. The second submission (run 2) is almost identical to the first submission, except that we replaced the squared versions of $\hat{k}^{LRD}_{3-6}$ and $\hat{k}_{\textit{audio}}$ with non-squared versions. Since all the submitted models are trained on both the provided training and development sets, we considered that KRR might provide better results if we choose a lower regularization parameter. Hence, the third submission (run 3) is also similar to the first submission, the only difference being that we changed the regularization parameter of KRR from $10^{-3}$ to $10^{-4}$.

\subsection{Results on the 2018 ADI Test Set}

\begin{table*}[!t]
\center
\begin{tabular}{|l|c|c|}
\hline
\bf System          & \bf Accuracy          & \bf Macro-$\mathbf{F_1}$ \\ 
\hline
\hline
Random Baseline     & -                     & $19.95\%$ \\
\hline
Run 1               & $\mathbf{58.65\%}$    & $\mathbf{58.92\%}$ \\
Run 2               & $54.50\%$             & $54.91\%$ \\
Run 3               & $58.36\%$             & $58.55\%$ \\
\hline
Run 4 (post-competition)    & $62.15\%$             & $62.02\%$ \\
Run 5 (post-competition)    & $\mathbf{62.22\%}$             & $\mathbf{62.28\%}$ \\
\hline
\end{tabular}
\caption{Results on the test set of the 2018 ADI Shared Task (closed training) of our multiple kernel learning method based on KRR versus a random baseline. The best results during and after the competition are highlighted in bold.}
\label{tab_results_ADI}
\vspace*{-0.5em}
\end{table*}

Table~\ref{tab_results_ADI} presents our results for the Arabic Dialect Identification Closed Shared Task of the 2018 VarDial Evaluation Campaign. Among the submitted systems, the best performance is obtained when the KRR regularization parameter is set to $10^{-3}$ and the kernel combination includes squared versions of $\hat{k}^{LRD}_{3-6}$ and $\hat{k}_{\textit{audio}}$. The submitted systems were ranked by their macro-$F_1$ scores, and among the $6$ participants, our best model obtained the first place with a macro-$F_1$ score of $58.92\%$. Remarkably, the statistical significance tests performed by the organizers indicate that our best system is significantly better than the system that ranked on the second place with a macro-$F_1$ score of $57.59\%$.

\begin{figure*}
\centering
\includegraphics[width=0.65\textwidth]{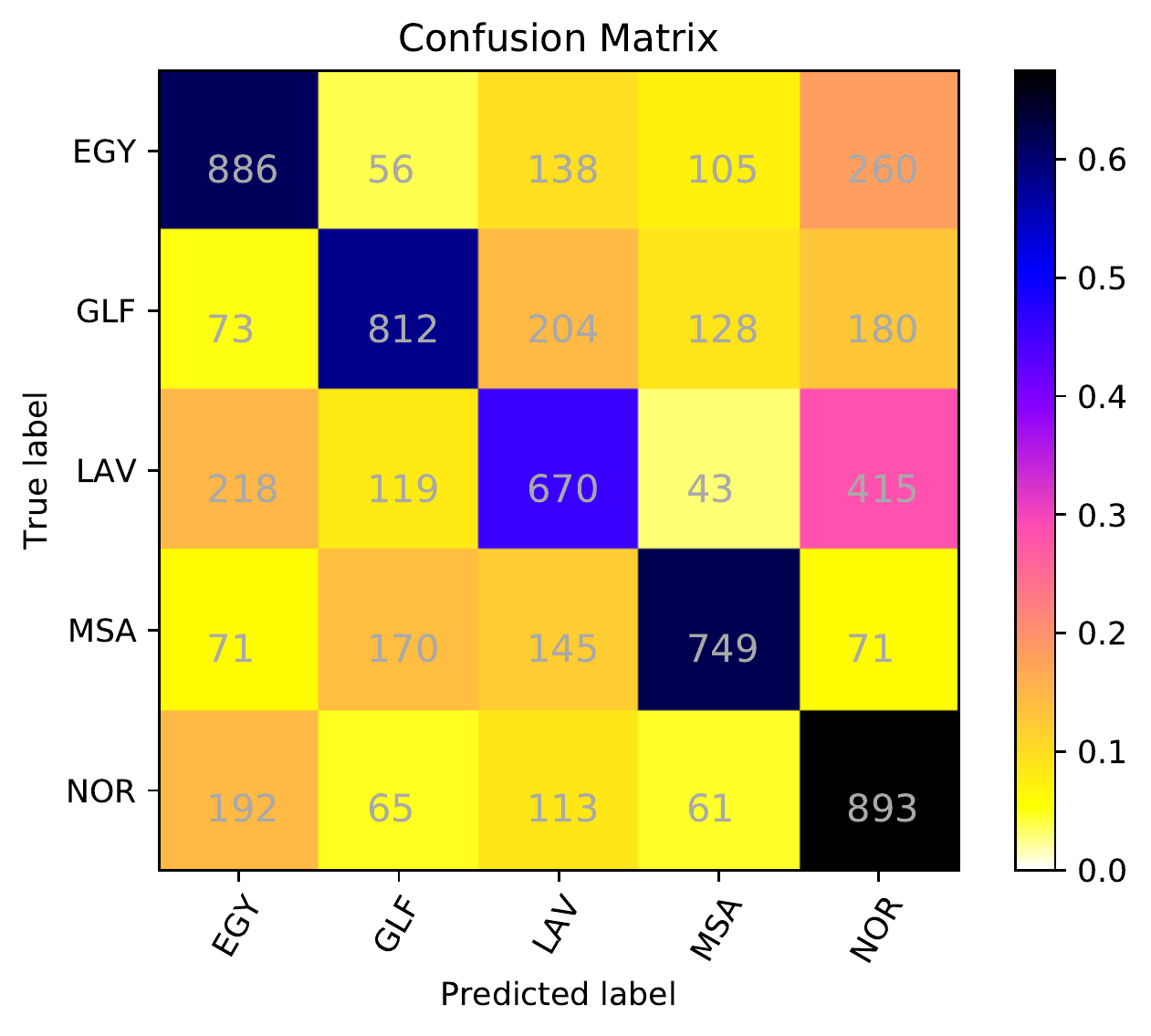}
\vspace*{-0.5em}
\caption{Confusion matrix (on the test set) of KRR based on the sum of $\hat{k}_{speech}$, $\hat{k}_{\textit{audio}}$ and $\hat{k}_{phonetic}$ (run 1). Best viewed in color.}
\label{fig_ADI_confusion_best}
\vspace*{-0.5em}
\end{figure*}

The confusion matrix for our best model is presented in Figure~\ref{fig_ADI_confusion_best}. The confusion matrix reveals that our system wrongly predicted more than $400$ examples of the Levantine dialect as part of the Maghrebi dialect. Furthermore, it has some difficulties in distinguishing the Levantine dialect from the Egyptian dialect on one hand, and the Egyptian dialect from the Maghrebi dialect on the other hand. Overall, the results look good, as the main diagonal scores dominate the other matrix components.

After the competition ended, the organizers released a new set of audio embeddings that are computed on both the training and the development sets. By replacing the audio embeddings released during the competition with the new embeddings (run 4), we obtained results that are more than $3\%$ better than out top system (run 1) during the competition. We also tried to include both embeddings based on audio features (run 5) into our multiple kernel learning approach, resulting in a slight improvement. The results of our post-competition runs 4 and 5 are reported in Table~\ref{tab_results_ADI}. Our best post-competition results are an accuracy of $62.22\%$ and a macro-$F_1$ score of $62.28\%$, obtained by run 5.

\section{Conclusion}
\label{sec_Conclusion}

In this paper, we presented an approach based on learning with multiple kernels for the ADI Shared Tasks of the 2018 VarDial Evaluation Campaign~\cite{vardial2018report}. Our approach attained very good results, as our team (UnibucKernel) ranked on the first place in the 2018 ADI Shared Task. The fact that we obtained the first place for the second year in a row indicates that our multiple kernel learning method did not reach the top performance simply by chance. We therefore conclude that the approach of combining multiple kernels based on different kinds of input, i.e. audio recordings, speech transcripts and phonetic transcripts, provides state-of-the-art performance in Arabic dialect identification. Since most of our kernels are based on character $p$-grams, we can also conclude that character $p$-grams represent the best feature set for the ADI task.

\bibliography{vardial2018}
\bibliographystyle{acl}

\end{document}